\documentclass[letterpaper, 10 pt, conference]{ieeeconf}

\IEEEoverridecommandlockouts
\usepackage{amsmath,amssymb,amsfonts}
\usepackage{algorithmic}
\usepackage{graphicx}
\usepackage{textcomp}
\usepackage{xcolor,soul}
\usepackage{url}
\usepackage{stfloats}
\usepackage{hyperref}
\usepackage{inputenc}
\usepackage{multirow, rotating}

\usepackage{subcaption}

\title{\LARGE \bf
Surrogate Model-Based Near-Optimal Gain Selection for Approach-Angle-Constrained Two-Phase Pure Proportional Navigation}

\author{
Abhigyan Roy$^{1}$%
\thanks{$^{1}$Student, Department of Aerospace Engineering,
Indian Institute of Technology Madras, Chennai, India.
Email: abhigyanroy191@gmail.com 
}%
\and
Shreeya Padte$^{2}$%
\thanks{$^{2}$ Research Associate, InspeCity, Mumbai, India.}
\and
Abel Viji George$^{1}$%
\and
A. Vivek$^{3}$%
\thanks{$^{3}$ Research Associate, Department of Aerospace Engineering,
Indian Institute of Technology Madras, Chennai, India.}
\and
Satadal Ghosh$^{4}$
\thanks{$^{4}$Associate Professor, Department of Aerospace Engineering,
Indian Institute of Technology Madras, Chennai, India.
Email: satadal@iitm.ac.in.}%
}

\begin{document}

\maketitle
\thispagestyle{empty}
\pagestyle{empty}

\begin{abstract}
In guidance literature, Pure Proportional Navigation (PPN) guidance is widely used for aerodynamically driven vehicles. A two-phase extension of PPN (2pPPN), which uses different navigation gains for an orientation phase and a final phase, has been presented to achieve any desired approach angle within an angular half-space. Recent studies show that the orientation phase can be realized through multiple feasible trajectories, creating an opportunity to select navigation gains that minimize overall guidance effort. This paper addresses the problem of near-optimal gain selection for given initial and desired terminal engagement geometries. Two optimization problems are considered: i) determination of the optimal orientation-phase gain for a specified final-phase gain, and ii) simultaneously determining the optimal gain pair for both phases that minimizes the total guidance effort. Determining the optimal gains analytically for arbitrary engagement geometries is intractable. Numerical simulations further reveal that these optimal gains vary smoothly with respect to the engagement conditions. Exploiting this property, a neural network (NN)–based regression model is developed in this paper to learn the nonlinear mapping between optimal gains and initial and desired terminal engagement geometries. The trained NN serves as a computationally efficient surrogate for generating the optimal gains manifold, enabling near-optimal realization of 2pPPN guidance. Numerical simulation studies demonstrate that the developed NN-based architecture predicts optimal gains with high accuracy, achieving very high (close to $0.9$) value of coefficient of determination.
\end{abstract}


\section{Introduction}
Besides simple point-to-point guidance, modern autonomous vehicle missions often demand precise control over the vehicle's state during different phases of the mission, especially at the terminal phase. In particular, controlling the terminal approach angle is necessary across diverse applications, such as, ensuring the perpendicularity required for effective autonomous industrial inspections, guiding UAVs safely toward recovery nets in restricted spaces, performing precise maneuvers such as motion through windows or narrow openings, docking to a charging station, and optimizing the impact geometry of loitering munitions to maximize attack efficiency. At the same time, autonomous vehicle operations are typically constrained by limited on-board energy, necessitating the development of guidance strategies that achieve terminal objectives with minimal guidance effort. 
To this end, this paper presents a guidance strategy that can compute near-optimal navigation gains for proportional navigation-based terminal angle control using a neural-network-based surrogate model.

Terminal angle-constrained interception has been widely studied using optimal control and nonlinear guidance approaches. An early contribution to optimal guidance law (OGL) was presented in \cite{Idan1995}, where the desired terminal angle was imposed as a terminal constraint while the approach time and control energy are minimized. Ryoo et al. \cite{Ryoo2005} derived a closed-form expression of an OGL dependent on time-to-go using the linearized engagement geometry. Later, in \cite{Ryoo2006}, time-to-go-based weights were added in the objective function to provide enhanced robustness to external disturbances. Nonlinear guidance methods based on Sliding Mode Control (SMC) allowed to account the nonlinear engagement dynamics. Representative works include \cite{Shima2011}, which enabled achieving a wide range of terminal approach angles; \cite{Kumar2012}, which guaranteed finite time convergence even in the presence of large initial heading errors; and \cite{He2019}, which improved the performance by integrating the reaching and sliding phases of SMC. In addition, the differential games framework approach have been used to derive nonlinear \cite{Bardhan2015} and linear \cite{Shaferman2017} feedback laws for achieving desired terminal geometries. Furthermore, trajectory-shaping techniques have been explored, where the look-angle is shaped as polynomial function of time to satisfy the terminal angle constraint \cite{Kang2019}. OGLs rely on linearized engagement kinematics and therefore are mostly valid near collision conditions, whereas SMC-based techniques require careful parameter tuning and depend on state variables that are difficult to measure.

In contrast, the Proportional Navigation (PN) guidance and its variants are widely used due to their simple command structure, minimal measurement requirements, optimal guidance effort under certain conditions \cite{Zarchan2012}, and robustness.
Pure Proportional Navigation (PPN) guidance with a time-varying bias term to achieve desired terminal angles has been proposed in several studies, including \cite{Kim1998}, \cite{Lee2013}, and \cite{Erer2016}. In these approaches, the PN component ensures target interception, while the additional bias term enables satisfaction of the desired approach angle. Alternatively, \cite{Ratnoo2008} introduced a two-phase PPN guidance scheme (2pPPN), where different gains were used in the Orientation and standard PPN phases to achieve any desired terminal angle within a half angular spectrum while also satisfying the bounded lateral acceleration requirement. This framework was later extended to achieve all possible approach directions against a stationary point in \cite{Ghosh2017}. 
However, these approaches restrict the relative angular configurations achievable at the end of the Orientation phase.
To address this limitation, an angular-plane trajectory shaping method for approach angle control was proposed in \cite{Vivek2025}, enabling the pursuer to achieve a broader set of relative angular configurations at the end of the Orientation phase. 

For a given set of initial and desired terminal engagement conditions, this formulation opens up the possibility of selecting the navigation gains in the two phases to minimize the overall guidance effort. However, determining such optimal gains for arbitrary engagement geometries is analytically intractable and often requires extensive numerical exploration of the admissible gain-space. This motivates the development of a neural network(NN)-based model capable of learning the mapping between engagement geometry and near-optimal gains, thereby enabling a near-optimal 2pPPN guidance implementation. 
Recently, learning-based approaches such as deep reinforcement learning  \cite{Lee2023} and predictor–corrector architectures using neural networks \cite{Luo2025} have been proposed to achieve desired terminal angles.

While effective, these approaches demand significant computational resources for training as well as online inference. In contrast, the proposed method leverages the inherent advantages of PPN guidance command and develops a lightweight NN-based regression model to predict the optimal navigation gains required to achieve a desired approach angle over the half-angular spectrum. This NN model acts as a computationally efficient surrogate for the optimal gain manifold, enabling near-optimal performance of 2pPPN with minimal computational overhead.

The remainder of this paper is organized as follows. Section~\ref{sec:background} provides the necessary background on 2pPPN guidance and presents the problem statement. Section~\ref{sec:methodology} describes the dataset generation procedure along the neural network architecture and training process. Section~\ref{sec:results_and_discussions} presents and discusses the results obtained using the trained network. Finally, Section~\ref{sec:conclusion} concludes the paper.

\section{Background and Problem Statement} \label{sec:background}
\subsection{Engagement Geometry and Kinematics} \label{subsec:engagement_geometry}
\begin{figure}[b]
    \centering
    \includegraphics[width=0.65\linewidth]{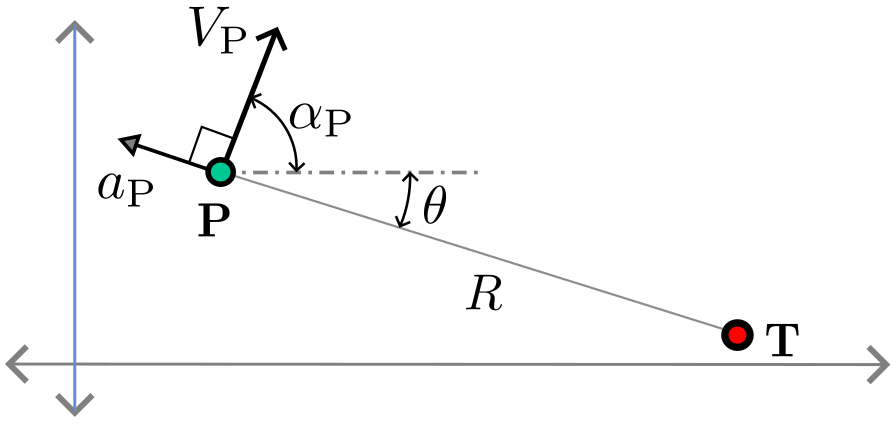}
    \caption{Engagement Geometry}
    \label{fig:engage_geom}
\end{figure}

Consider a planar engagement between a pursuing UAV (pursuer) and a stationary target as shown in Fig \ref{fig:engage_geom}. Let $R$ denote the line-of-sight (LOS) range, $\theta$ be the LOS angle measured with respect to (w.r.t.) an inertial reference, and $\alpha_{\mathrm{P}}$ be the pursuer heading angle. The pursuer moves with constant speed $V_P$ and is commanded by a lateral acceleration $a_{\mathrm{P}}$, which is applied normal to its velocity vector. In this paper, subscripts $\mathrm{0}$ and $\mathrm{f}$ are used to denote the initial and final values, respectively, and the desired value is represented with a superscript $\mathrm{d}$. 
The equations governing the kinematics of this engagement are given by,
\begin{align} 
    \dot{R} &= -V_P \cos(\alpha_{\mathrm{P}} - \theta); \label{eqn:los_range_rate} \\ 
    \dot{\theta} &= -\frac{V_P}{R} \sin(\alpha_{\mathrm{P}} - \theta);\:\:
    \dot{\alpha}_P = a_{\mathrm{P}}/V_P \label{eqn:los_rate_pursuer_turnrate}
\end{align}
In this paper, the 2pPPN guidance framework is employed to achieve the desired terminal approach angle. The following sections briefly present the necessary preliminaries of PPN and 2pPPN.

\subsection{Pure Proportional Navigation (PPN)} \label{subsec:PPN_basics}
For a constant-speed pursuer following PPN guidance, the lateral acceleration command is given by
\begin{align}
    a_{\mathrm{P}} = N V_P \dot{\theta}, \label{eq:ppn_command} 
\end{align} 
where $N$ is the navigation gain. 
Substituting \eqref{eq:ppn_command} into \eqref{eqn:los_rate_pursuer_turnrate} and integrating w.r.t time yields $\alpha_{\mathrm{P}} - \alpha_{\mathrm{P}_\mathrm{0}} = N (\theta - \theta_0)$. For interception with a stationary target, the collision condition is $\alpha_{\mathrm{P}_\mathrm{f}} = \theta_f$.  Therefore, for a given initial engagement geometry ($\theta_0,\alpha_{\mathrm{P}_\mathrm{0}}$) and navigation gain $N$, the achieved terminal angle can be obtained as 
\begin{align} 
    \alpha_{\mathrm{P}_\mathrm{f}} = (N\theta_0 - \alpha_{\mathrm{P}_\mathrm{0}})/(N - 1). \label{eq:alpha_Pf_ppn} 
\end{align}
To ensure bounded terminal lateral acceleration against stationary targets, the navigation gain must satisfy $N\geq 2$ \cite{Guelman1971}. Consequently, for $N\in[2,\infty)$, the achievable terminal angles using standard single phase PPN are restricted to 
\begin{align} 
    \alpha_{\mathrm{P}_\mathrm{f}} \in [2\theta_0 - \alpha_{\mathrm{P}_\mathrm{0}}, \theta_0). \label{eq:ppn_band} 
\end{align}
However, for a stationary target located on ground, $\alpha_{\mathrm{P}_\mathrm{0}} > \theta_0$. Hence, from \eqref{eq:ppn_band}, the set of achievable terminal angles is only a small subset of the angular half-space $[-\pi + \theta_0, \theta_0)$. Similarly, for $\alpha_{\mathrm{P}_\mathrm{0}} < \theta_0$ case, achievable terminal angles lie in a subset of $(\theta_0, \pi + \theta_0]$.

\subsection{Two-Phase Pure Proportional Navigation (2pPPN)} \label{subsec:2pPPN_basics}
From the discussion in Section~\ref{subsec:PPN_basics}, it is evident that a significant portion of the feasible approach angle-space $[-\pi+\theta_0,\theta_0)$ cannot be realized using a standard single phase PN guidance. Clearly, the reason for the same is that for the remaining portion of terminal angle spectrum, $N=(\alpha_{\mathrm{P}_\mathrm{f}}^{\mathrm{d}} - \alpha_{\mathrm{P}_\mathrm{0}})/(\alpha_{\mathrm{P}_\mathrm{f}}^{\mathrm{d}} - \theta_0)$ is less than $2$. To overcome this limitation, a 2pPPN guidance strategy was presented that consisted of Orientation and Final phases (\cite{Ratnoo2008} and \cite{Ghosh2017}). The Orientation phase trajectory was designed such that at the end of it $\alpha_{\mathrm{P}}=\theta_0$ and $\theta=\theta_0-\pi/2$ were achieved. It was shown that at some point $(\alpha_{\mathrm{P}},\theta)$ on the Orientation phase trajectory, required navigation gain $\left(N_{\mathrm{req}}\triangleq(\alpha_{\mathrm{P}_\mathrm{f}}^{\mathrm{d}} - \alpha_{\mathrm{P}})/(\alpha_{\mathrm{P}_\mathrm{f}}^{\mathrm{d}} - \theta)\right)$ for the Final phase would be greater than 2. Thus, the navigation gains for the 2pPPN guidance were given as,
\begin{align}
    N =
    \begin{cases}
        N_{\mathrm{ori}}=(2/\pi)|\alpha_{\mathrm{P}_\mathrm{0}}-\theta_0|  & \text{if }   N_{\mathrm{req}} < 2    \\
        N_{\mathrm{f}}=N_{\mathrm{req}}     & \text{if }   N_{\mathrm{req}} \geq 2
    \end{cases}
    \label{eq:2pppn_gain}
\end{align}

\begin{figure}
    \centering
    \includegraphics[width=0.55\linewidth]{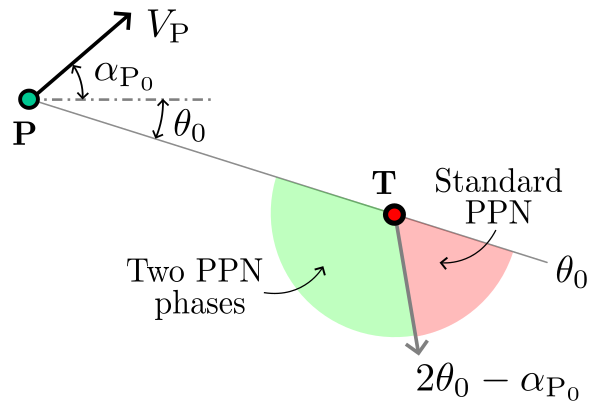}
    \caption{Angular half-space covered by 2pPPN}
    \label{fig:2pPPN_halfspace}
\end{figure}

\subsection{Problem Statement} \label{subsec:problem_statement}
Note from Section~\ref{subsec:2pPPN_basics} that while the notion of extending a standard PPN guidance to a two-phase PPN guidance facilitated expanding the set of achievable approach angles to the entire angular half-space that depends on $\theta_0$ and $\alpha_{\mathrm{P}_\mathrm{0}}$, the choice of end configuration of Orientation phase trajectory, as presented in \cite{Ratnoo2008} and \cite{Ghosh2017}, in terms of $\alpha_{\mathrm{P}}$ and $\theta$ was very specific. There may be countably many number of possibilities of such selection of end configurations of Orientation phase trajectory, each one of which demands a certain magnitude of overall guidance effort. This opens up an opportunity to optimize the selection of end configuration of Orientation phase trajectory or equivalently selection of ($N_{\mathrm{ori}}, N_{\mathrm{f}}$) such that the overall guidance effort 
\begin{align}
    J(N_{\mathrm{ori}}, N_{\mathrm{f}}, \alpha_{\mathrm{P}_\mathrm{0}}, \alpha^{\mathrm{d}}_{\mathrm{P}_\mathrm{f}}) = \int_0^{t_f} a_{\mathrm{P}}^{2} \, dt \label{eqn:objective_fn1}
\end{align}
is minimized. 

Accordingly, with the objective of minimizing the total guidance effort $J$, this paper addresses the following problems.
\begin{enumerate}[label=(\Alph*)]
    \item For given engagement conditions $\alpha_{\mathrm{P}_\mathrm{0}}, \alpha^{\mathrm{d}}_{\mathrm{P}_\mathrm{f}}$, and $N_{\mathrm{f}}$, determine the near-optimal Orientation phase navigation gain $N_{\mathrm{ori}}^{*}$, subject to bounds on $N_{\mathrm{ori}}$ and $N_{\mathrm{f}}$. 
    \item For given engagement conditions $\alpha_{\mathrm{P}_\mathrm{0}}$ and $\alpha^{\mathrm{d}}_{\mathrm{P}_\mathrm{f}}$, determine the near-optimal navigation gain pair $(N_{\mathrm{ori}}^{*},N_{\mathrm{f}}^{*})$, subject to bounds on $N_{\mathrm{ori}}$ and $N_{\mathrm{f}}$.
\end{enumerate}

\section{Methodology} \label{sec:methodology}
In order to solve the optimization problem defined in Section~\ref{subsec:problem_statement}, a data-driven methodology is developed to learn the relationship between the 2pPPN gains and the initial and desired terminal engagement conditions within the feasible engagement region. First, feasible bounds on the navigation gains are established. Next, the required training dataset is generated through numerical simulations. Finally, a neural network is trained to learn and interpolate the optimal gain mappings for 2pPPN.

\subsection{Navigation Gain Bounds} \label{subsec:navgain_bounds}
\begin{figure}[h]
    \centering
    \includegraphics[width=0.6\linewidth]{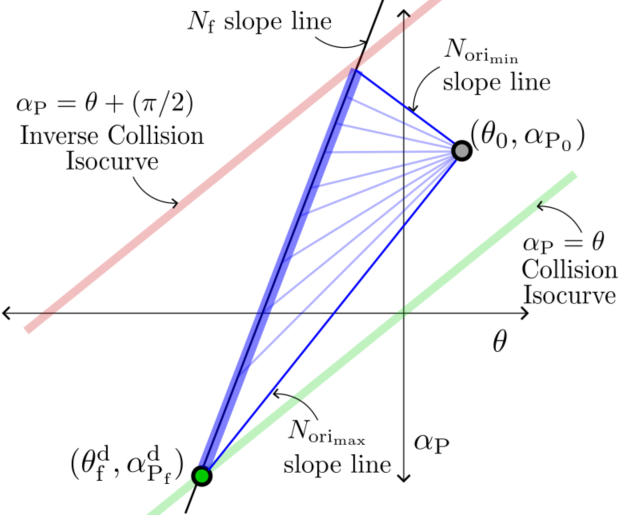}
    \caption{Possible orientation phase gains}
    \label{fig:gain_bound}
\end{figure}

The selection of $N_{\mathrm{ori}}$ in \cite{Ratnoo2008} and \cite{Ghosh2017} restricts the engagement to attain a specific angular configuration $\mathrm{P}_{\mathrm{ori}_\mathrm{f}}\triangleq(\theta_{\mathrm{ori}_{\mathrm{f}}},\alpha_{\mathrm{P}_{\mathrm{ori,\,f}}})$ at the end of the Orientation phase. However, analysis in $(\theta,\alpha_{\mathrm{P}})$ plane (\cite{Vivek2025}) shows that $N_{\mathrm{ori}}$ can lie within a feasible interval $N_{\mathrm{ori}}\in [N_{\mathrm{ori}_{\min}},N_{\mathrm{ori}_{\max}}]$ as shown in Fig.~\ref{fig:gain_bound}. In this angular plane, the slope of the line joining two $(\theta,\alpha_{\mathrm{P}})$ points represents the corresponding PPN navigation gain. Hence, reaching the desired terminal configuration $\mathrm{P}_{\mathrm{f}}^{\mathrm{d}}\triangleq(\alpha_{\mathrm{P}_{\mathrm{f}}}^{\mathrm{d}},\alpha_{\mathrm{P}_{\mathrm{f}}}^{\mathrm{d}})$ with bounded lateral acceleration requires approaching $\mathrm{P}_{\mathrm{f}}$ along a line with a slope $\geq2$ (i.e., $N_{\mathrm{f}}\geq2$). Thus, the Orientation phase drives the engagement configuration to this line on $(\theta,\alpha_{\mathrm{P}})$ plane. These constraints restrict the choice of $N_{\mathrm{f}}$ and $N_{\mathrm{ori}}$. 

In this paper, $N_{\mathrm{f}}\in[2,5]$ is considered. The lower limit arises from the bounded $a_{\mathrm{P}}$ requirement \cite{Guelman1971}, while the upper limit stems from practical autopilot and actuator limitations \cite{Zarchan2012}. The bounds on $N_{\mathrm{ori}}$ can be obtained geometrically from the admissible location of $\mathrm{P}_{\mathrm{ori}_{f}}$ on $N=N_{\mathbf{f}}$ line. i.e., $\mathrm{P}_{\mathrm{ori}_{f}}$ should lie on the segment of $N=N_{\mathrm{f}}$ line bounded by the collision isocurve $\alpha_{\mathrm{P}}=\theta$ and the inverse collision isocurve $\alpha_{\mathrm{P}}=\theta+\pi$.  
For more details on 2pPPN guidance based on phase plane trajectory shaping, refer to Section~II.C in \cite{Vivek2025}. 

As shown in Fig.~\ref{fig:gain_bound}, the upper bound $N_{\mathrm{ori}_{\max}}$ corresponds to the slope of the line connecting $(\theta_0,\alpha_{\mathrm{P}_\mathrm{0}})$ and the intersection of $\alpha_{\mathrm{P}}=\theta$ with $N=N_{\mathrm{f}}$ line, given by 
\begin{align}
    N_{\mathrm{ori}_{\max}} = (\alpha_{\mathrm{P}_\mathrm{f}}^{\mathrm{d}} - \alpha_{\mathrm{P}_\mathrm{0}})/(\alpha_{\mathrm{P}_\mathrm{f}}^{\mathrm{d}} - \theta_0), \label{eqn:max_N_ori}
\end{align}
Similarly, the lower bound $N_{\mathrm{ori}_{\min}}$ is obtained from the slope of the line connecting $(\theta_0,\alpha_{\mathrm{P}_\mathrm{0}})$ and the intersection of $\alpha_{\mathrm{P}}=\theta+\pi$ and $N=N_{\mathrm{f}}$ line, which gives
\begin{align}
    N_{\mathrm{ori}_{\min}}^{'}=\frac{
        \alpha_{\mathrm{P}_\mathrm{f}}^{\mathrm{d}} - \alpha_{\mathrm{P}_\mathrm{0}}
        + \pi N_{\mathrm{f}} /(N_{\mathrm{f}} - 1)}{\alpha_{\mathrm{P}_\mathrm{f}}^{\mathrm{d}} - \theta_0 + \pi/(N_{\mathrm{f}} - 1) }. \label{eqn:min_N_ori}
\end{align}
From Fig.~\ref{fig:gain_bound}, it can be observed that $N_{\mathrm{ori}}$ may take negative values. However, selecting $N_{\mathrm{ori}}<<0$ may cause the pursuer to initially turn away from the target, potentially resulting in an undesirably longer trajectory and increase the engagement time. Hence, in this paper, the lower bound is taken as $N_{\mathrm{ori}_{\min}}=\max{(N_{\mathrm{ori}_{\min}}^{'},-2)}$.  

\subsection{Dataset Generation for Feasible 2pPPN Cases}\label{sec:dataset_construct}
Without loss of generality, this paper assumes $\theta_{0}=0^{\circ}$ for dataset generation during neural network training. Additionally, due to symmetry, only cases where $\alpha_{\mathrm{P}_{0}}>\theta_{0}$ are considered. Since the objective involves minimizing $N_{\mathrm{ori}}$, only engagement conditions that require both PPN phases are included, i.e., $\alpha_{\mathrm{P}_{\mathrm{f}}}^{\mathrm{d}}<2\,\theta_{0}-\alpha_{\mathrm{P}_{0}}$. 
This corresponds to the portion of the angular half-space where the Orientation phase is necessary, as shown in Fig.~\ref{fig:2pPPN_halfspace}.

In addition to $N_{\mathrm{ori}}$ and $N_{\mathrm{f}}$ bounds given in Section~\ref{subsec:navgain_bounds}, constraint on $t_{\mathrm{f}}$ is also imposed to ensure practical mission feasibility. Without such a restriction, the optimization problem in \eqref{eqn:objective_fn1} may allow trivial solutions in which the pursuer executes extremely slow orientation phase maneuvers with arbitrarily small lateral acceleration while still satisfying the terminal angle condition asymptotically. Such trajectories are undesirable for practical UAV missions due to excessive engagement duration. Thus, the final time is restricted to $t_{\mathrm{f}}\leq t_{\max}$, where $t_{\max}$ is chosen to ensure at least some feasible gains exist within the admissible range for each $(\alpha_{\mathrm{P}_{0}}, \alpha_{\mathrm{P}_{\mathrm{f}}}^{\mathrm{d}})$.
\begin{figure*}
     \centering
     \begin{subfigure}[b]{\linewidth}
         \centering
         \includegraphics[width=0.7\textwidth]{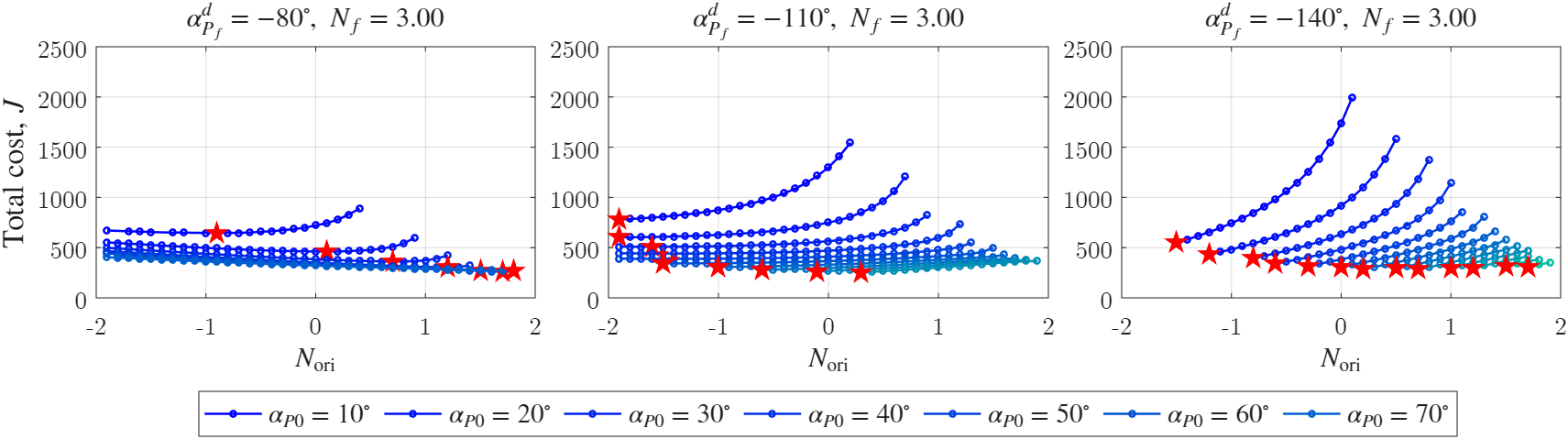}
         \caption{$N_{\mathrm{f}}=3.00$}
         \label{fig:Cost_Nori_Nf30}
     \end{subfigure}
     \begin{subfigure}[b]{\linewidth}
         \centering
         \includegraphics[width=0.7\textwidth]{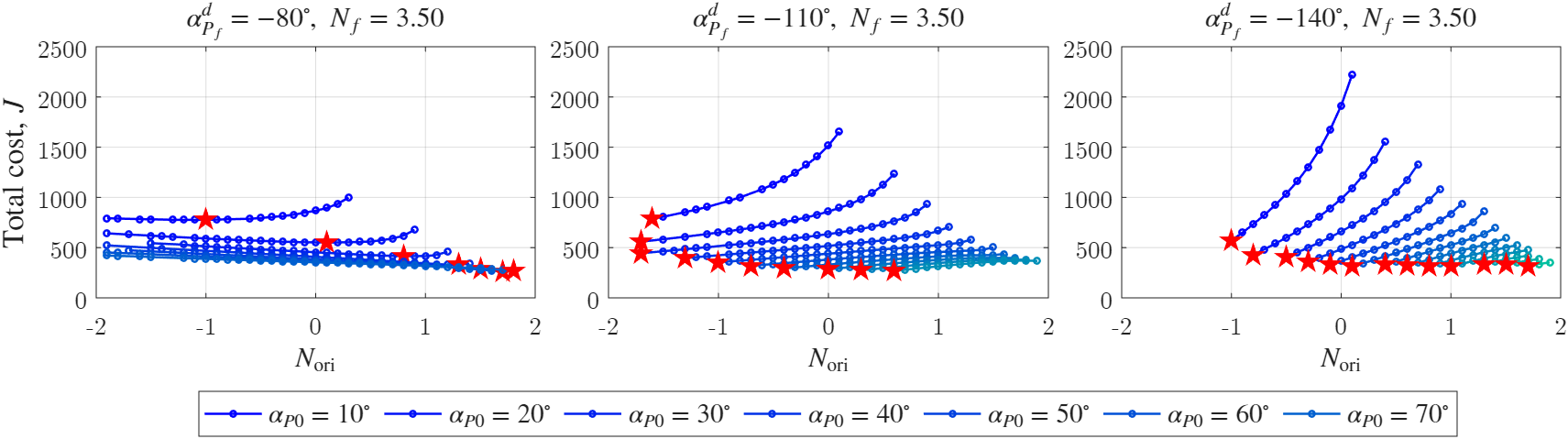}
         \caption{$N_{\mathrm{f}}=3.50$}
         \label{fig:Cost_Nori_Nf35}
     \end{subfigure}     
     \caption{Total cost variation v/s $N_{\mathrm{ori}}$ for different $\alpha_{\mathrm{P}_{0}}$, $\alpha_{\mathrm{P}_{\mathrm{f}}}^{\mathrm{d}}$, and $N_{\mathrm{f}}$ values.}
     \label{fig:cost_versus_Nori}
\end{figure*}

For each engagement geometry defined by $(\alpha_{\mathrm{P}_\mathrm{0}}, \alpha^{\mathrm{d}}_{\mathrm{P}_\mathrm{f}})$ pair, the optimal navigation gains are obtained by solving the following constrained optimization problems over the feasible gain region.
\begin{subequations} \label{eq:optimization_problems}
    \begin{align}
    N_{\mathrm{ori}}^* &= \underset{N_{\mathrm{ori}}}{\arg\min}\; 
    J(N_{\mathrm{ori}} | N_{\mathrm{f}}, \alpha_{\mathrm{P}_\mathrm{0}}, \alpha^{\mathrm{d}}_{\mathrm{P}_\mathrm{f}}), \label{eq:optimization_problem_a} \\
    (N_{\mathrm{f}}^*, N_{\mathrm{ori}}^*) &= \underset{N_{\mathrm{f}},\, N_{\mathrm{ori}}}{\arg\min} \; 
    J(N_{\mathrm{ori}}, N_{\mathrm{f}} | \alpha_{\mathrm{P}_\mathrm{0}}, \alpha^{\mathrm{d}}_{\mathrm{P}_\mathrm{f}}) \label{eq:optimization_problem_b}
    \end{align}
\end{subequations}
subject to the same constraints
\begin{align}
    t_f &\leq t_{\max}, \\ \label{eq:tc}
    2 &\leq N_{\mathrm{f}} \leq 5, \\
    N_{\mathrm{ori}}^{\min} &< N_{\mathrm{ori}} < N_{\mathrm{ori}}^{\max}.
\end{align}

The engagement conditions and navigation gains are varied over the following ranges. $\alpha^{\mathrm{d}}_{\mathrm{P}_\mathrm{f}}\in[-170^\circ,-10^\circ]$, $\alpha_{\mathrm{P}_\mathrm{0}}\in[10^\circ,170^\circ]$ in steps of $10^\circ$, $N_{\mathrm{f}}\in[2,5]$ and $N_{\mathrm{ori}}\in [N_{\mathrm{ori}_{\min}},N_{\mathrm{ori}_{\max}}]$ in steps of 0.1. Subsequently, the condition for 2pPPN $\alpha_{\mathrm{P}_{\mathrm{f}}}^{\mathrm{d}}<2\,\theta_{0}-\alpha_{\mathrm{P}_{0}}$ is applied, which resulted in a total of $94,857$ valid scenarios. For each parameter combination, the engagement dynamics are simulated to evaluate the corresponding cost and the optimal gains corresponding to the two optimization problems \ref{eq:optimization_problems} are obtained, giving us 4216 and 136 optimal cases, respectively. 

\begin{figure}
    \centering
    \includegraphics[width=0.8\linewidth]{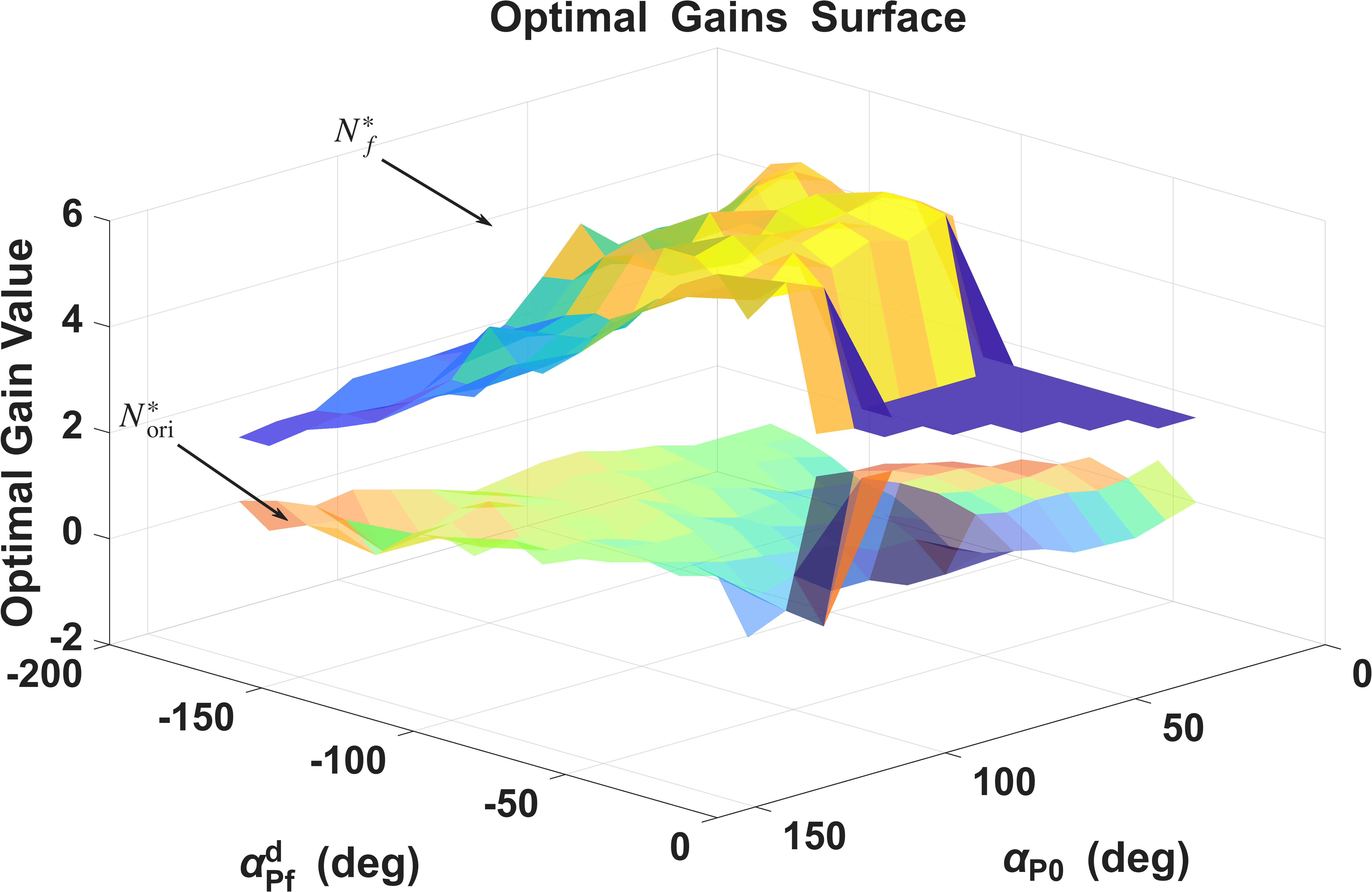}
    \caption{Optimal gain surfaces for different engagement geometry}
    \label{fig:opt_cost_surface_all}
\end{figure}

The variation of the total cost $J$ w.r.t to $N_{\mathrm{ori}}$ is illustrated in Fig.~\ref{fig:cost_versus_Nori} for selected engagement conditions and $N_{\mathrm{f}}$ values. A surface plot of optimal $N_{\mathrm{ori}}$ and $N_{\mathrm{f}}$ against the engagement conditions is shown in Fig.~\ref{fig:opt_cost_surface_all}. These figures indicate that the cost varies smoothly with both the navigation gains and the engagement parameters, and that the corresponding optimal gain manifold is smooth w.r.t the engagement conditions.

\subsection{Data-Driven Gain Interpolation}
The optimization procedure described in Section~\ref{sec:dataset_construct} produces a discrete set of optimal navigation gains corresponding to sampled engagement geometries. 
Numerical evaluation of the cost function $J(N_{\mathrm{ori}}, N_{\mathrm{f}}, \alpha_{\mathrm{P}_\mathrm{0}}, \alpha^{\mathrm{d}}_{\mathrm{P}_\mathrm{f}})$ given in \eqref{eqn:objective_fn1} over the feasible gain region shows that cost surface is smooth with well-defined minima across engagement configurations as shown in Fig.~\ref{fig:cost_versus_Nori}.
Consequently, the optimal gains $(N_{\mathrm{ori}}^{*}, N_{\mathrm{f}}^{*})$ form a smooth gain manifold over the engagement space defined by $(\alpha_{\mathrm{P}_\mathrm{0}}, \alpha^{\mathrm{d}}_{\mathrm{P}_\mathrm{f}})$. 
Although this optimal gain manifold can be computed offline through numerical optimization, obtaining the gains online would require repeated simulation, which is computationally impractical. To address this limitation and enable fast gain selection for online guidance command generation, a NN-based regression framework is employed to approximate the mapping between engagement geometry and optimal navigation gains and provide a computationally efficient surrogate for the optimal gain manifold.

Two NN configurations are considered corresponding to the two optimization formulations. Both models employ fully connected feedforward neural networks with tanh activation functions to capture the nonlinear dependence of the optimal orientation gains on the engagement configurations.
\subsubsection{Model A ($N_{\mathrm{ori}}^{*}$ Prediction-Problem \eqref{eq:optimization_problem_a})} \label{subsec:NNModel_A}
In this configuration, the terminal-phase navigation gain $N_{\mathrm{f}}$ is considered to be chosen by the guidance designer based on other performance considerations such as actuator limits or auto pilot lags. The neural network, as shown in Fig \ref{fig:regr_fw} is therefore required to predict the following relationship.
\begin{align}
    \mathcal{F}_{A} : (\alpha_{\mathrm{P}_\mathrm{0}}, \alpha^{\mathrm{d}}_{\mathrm{P}_\mathrm{f}}, N_{\mathrm{f}}) \to N_{\mathrm{ori}}^{*}
\end{align}
In this model, the final gain is treated as a design input. There are 3 inputs neurons, 2 hidden layers with 32 and 16 neurons, and one output neuron corresponding to $N_{\mathrm{ori}}^{*}$. 

\subsubsection{Model B ($(N_{\mathrm{ori}}^{*},N_{\mathrm{f}}^{*})$ pair Prediction-Problem \eqref{eq:optimization_problem_b})}  \label{subsec:NNModel_B}
\begin{figure}[h]
    \centering
    \includegraphics[width=0.7\linewidth]{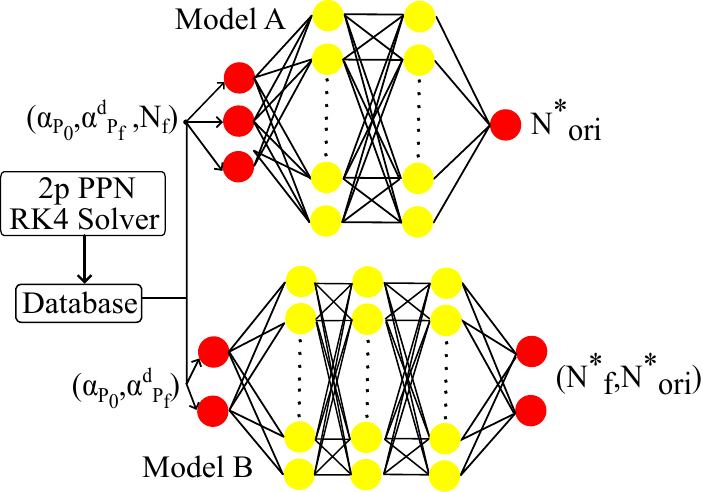}
    \caption{Data-driven Framework}
    \label{fig:regr_fw}
\end{figure}

In the second configuration, the neural network, as shown in Fig \ref{fig:regr_fw} directly predicts the optimal pair of navigation gains corresponding to the optimization problem~\eqref{eq:optimization_problem_b}. Therefore, the neural network predicts the following relationship.
\begin{align}
    \mathcal{F}_{B} : (\alpha_{\mathrm{P}_\mathrm{0}}, \alpha^{\mathrm{d}}_{\mathrm{P}_\mathrm{f}}) \to (N_{\mathrm{ori}}^{*},N_{\mathrm{f}}^{*})
\end{align}
This model has 2 inputs neurons corresponding to the engagement geometry, 3 hidden layers with 32, 64 and 16 neurons each, and 2 output neurons corresponding to $(N_{\mathrm{ori}}^{*},N_{\mathrm{f}}^{*})$. 

\subsubsection{Training Procedure} \label{subsec:NN_training}
The training dataset is generated from the full engagement sweep described previously in Section~\ref{sec:dataset_construct}, with $N_s$ being the number of optimal scenarios the model is being trained on  ($N_s$ = 4216 for Model A and $N_s$ = 136 for Model B).  

The dataset is then partitioned into training and testing subsets using an $80/20$ split to evaluate generalization performance.
Prior to training, all input and output variables are normalized to zero mean and unit variance to improve numerical conditioning. 

The networks are trained using the Adam optimizer, for its adaptive learning rate and fast convergence properties \cite{Kingma2014Adam}, with a learning rate of $10^{-3}$. Model A and Model B are trained for $20000$ and $40000$ epochs, respectively. The following Mean squared error(MSE)-based loss function is considered for both models. 
\begin{equation}
    \mathcal{L} = \frac{1}{N_s} \sum_{i=1}^{N_s}\left\| \hat{y}_i - y_i \right\|^2, \label{eqn:loss_function}
\end{equation}
where $y_{i}$ denotes the optimal navigation gains obtained from the numerical sweep and $\hat{y}_{i}$ denotes the network prediction.

Once trained, the NN provides a continuous approximation of the optimal gain manifold obtained from the numerical optimization procedure. At the beginning of the engagement, the initial engagement geometry and, where applicable, the final navigation gain $N_{\mathrm{f}}$ are supplied as inputs to the network, and the corresponding navigation gains are obtained through a single forward evaluation. This eliminates the need for repeated online optimization. 

\section{Results and Discussions} \label{sec:results_and_discussions}
The simulations are performed for a planar engagement scenario with $V_P = 50~\mathrm{m/s}$, a stationary target ($V_T = 0$), initial separation fixed at $R_0 = 2500~\mathrm{m}$, and the initial line-of-sight angle $\theta_0 = 0$. This workflow uses an AMD Ryzen 7 3.80 GHz computer to generate datasets in MATLAB, subsequently processed in Python 3.10, using NumPy 2.2, and PyTorch 2.7 to implement an NN regression framework.

The trained NNs learn the mappings for both models using the dataset of optimal gain pairs extracted from the sweep. To evaluate generalization, multiple engagement cases are selected randomly from the admissible domain in the test set. Table~\ref{tab:prediction} reports representative examples comparing NN predictions with ground-truth optimal gains obtained via direct minimization. We can see that we obtain sufficient accuracy in the NN predictions. But as we move towards the boundary cases of 2pPPN, the error in both models increases.  
\vspace{-10pt}
\begin{table}[h]
\caption{Neural Network Prediction Accuracy for Unseen Engagements}
\centering
\textbf{Model A: } $(\alpha_{\mathrm{P}_\mathrm{0}}, \alpha^{\mathrm{d}}_{\mathrm{P}_\mathrm{f}}, N_{\mathrm{f}}) \rightarrow N_{\mathrm{ori}}^*$\\[3pt]
\begin{tabular}{|c|c|c|c|c|}
\hline
$\alpha_{\mathrm{P}_\mathrm{0}}$ & $\alpha^{\mathrm{d}}_{\mathrm{P}_\mathrm{f}}$ & $N_{\mathrm{f}}$
& True $N_{\mathrm{ori}}^*$
& Predicted $N_{\mathrm{ori}}^*$ \\
\hline
$25^\circ$ & $-30^\circ$  & $2.00$ & $1.70$ & $1.70$ \\
\hline
$25^\circ$ & $-90^\circ$  & $2.00$ & $-0.20$  & $-0.34$  \\
\hline
$25^\circ$ & $-150^\circ$ & $3.20$ & $-0.60$  & $-1.13$  \\
\hline
\end{tabular}

\vspace{6pt}
\textbf{Model B: } $(\alpha_{\mathrm{P}_\mathrm{0}}, \alpha^{\mathrm{d}}_{\mathrm{P}_\mathrm{f}}) \rightarrow (N_{\mathrm{f}}^*, N_{\mathrm{ori}}^*)$\\[3pt]
\begin{tabular}{|c|c|c|c|}
\hline
$\alpha_{\mathrm{P}_\mathrm{0}}$ & $\alpha^{\mathrm{d}}_{\mathrm{P}_\mathrm{f}}$
& True $(N_{\mathrm{f}}^*, N_{\mathrm{ori}}^*)$
& Predicted $(N_{\mathrm{f}}^*, N_{\mathrm{ori}}^*)$ \\
\hline
$25^\circ$ & $-30^\circ$ & $(2.00,\ 1.71)$ & $(2.00,\ 1.60)$ \\
\hline
$25^\circ$ & $-90^\circ$ & $(2.00,\ -0.20)$    & $(2.02,\ -0.31)$    \\
\hline
$25^\circ$ & $-150^\circ$ & $(3.20,\ {-}0.60)$ & $(4.99,\ -0.41)$ \\
\hline
\end{tabular}
\label{tab:prediction}
\end{table}

\begin{figure}[h]
    \centering
    \includegraphics[width=0.75\linewidth]{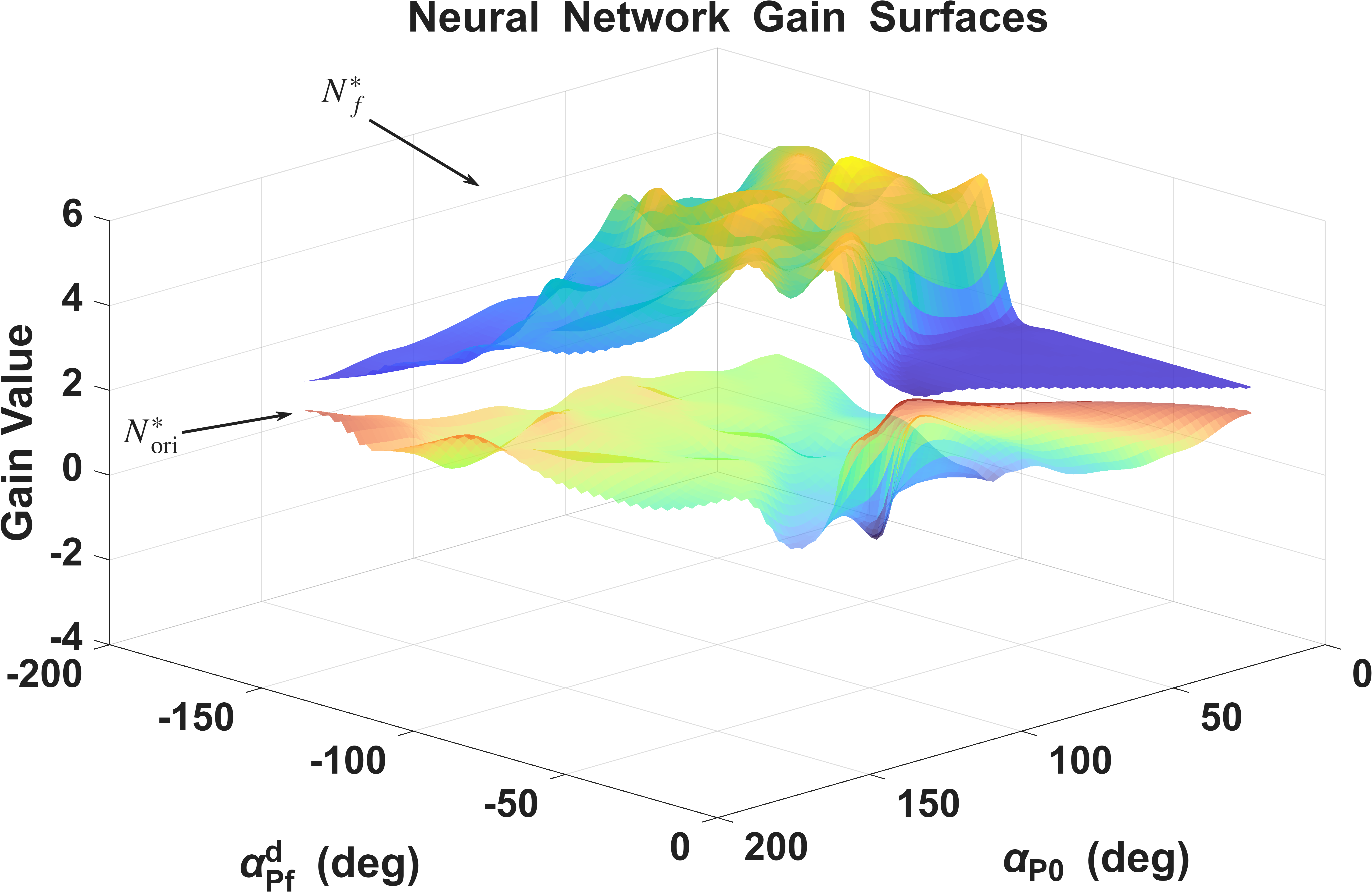}
    \caption{NN Predicted Optimal gain surfaces for different engagement geometry}
    \label{fig:NN_opt_cost_surface_all}
\end{figure}

Regression accuracy is evaluated on a validation set using the coefficient of determination $R^2$. Model~A, which maps $(\alpha_{\mathrm{P}_\mathrm{0}}, \alpha^{\mathrm{d}}_{\mathrm{P}_\mathrm{f}}, N_{\mathrm{f}})$ to $N_{\mathrm{ori}}^*$, achieves $R^2 = 0.932$, with predictions tightly clustered about the ideal line across the full output range. Model~B, predicting $(N_{\mathrm{f}}^*, N_{\mathrm{ori}}^*)$ jointly from $(\alpha_{\mathrm{P}_\mathrm{0}}, \alpha^{\mathrm{d}}_{\mathrm{P}_\mathrm{f}})$ alone, yields $R^2 = 0.880$ and $R^2 = 0.901$ for the two outputs, respectively, with the largest deviations occurring at the extremes of the $N_{\mathrm{f}}^*$ range where the training data is sparser. Considering the $R^2$ value of of both models, we can see that the higher accuracy of Model~A is consistent with the additional conditioning on $N_{\mathrm{f}}$ in the training process, which reduces the ambiguity in the regression target.
\begin{figure}
    \centering
    \includegraphics[width=\linewidth]{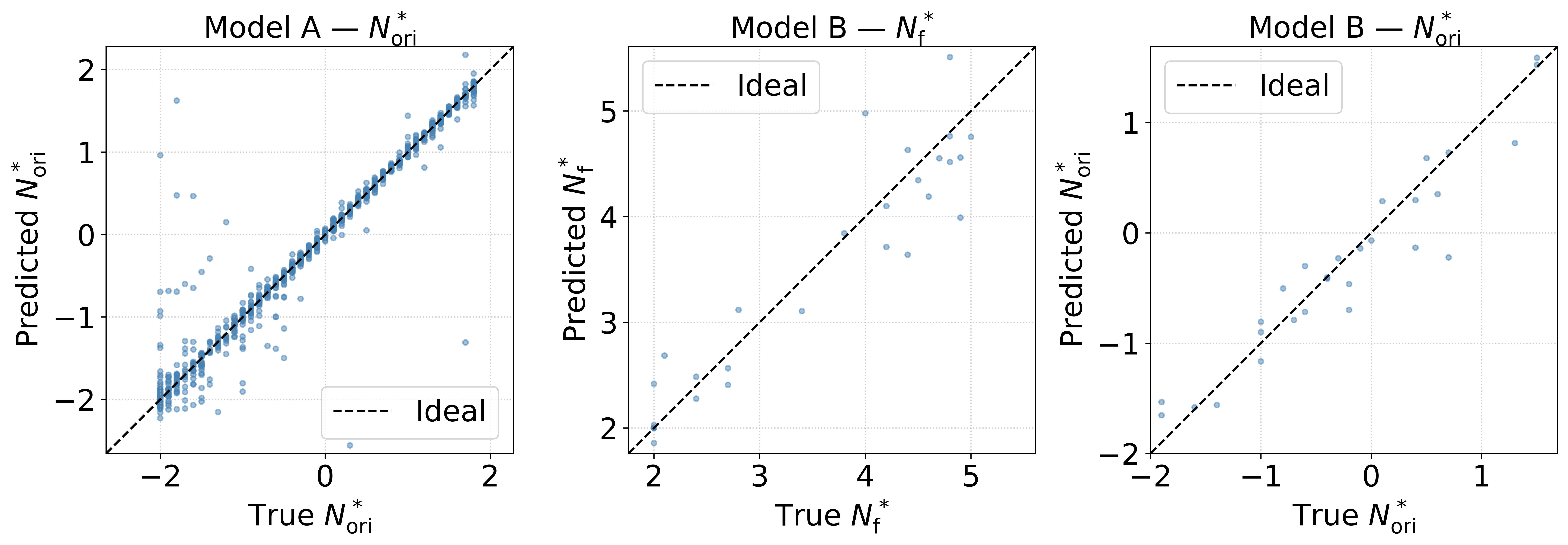}
    \caption{Predicted vs. True Optimal Gains (Test Set)}
    \label{fig:placeholder}
\end{figure}
Figure~\ref{fig:NN_opt_cost_surface_all} shows the corresponding NN predictions of the optimal gain manifolds obtained from the parameter sweep, as a function of $(\alpha_{\mathrm{P}_\mathrm{0}}, \alpha^{\mathrm{d}}_{\mathrm{P}_\mathrm{f}})$. The NN-predicted surfaces exhibit the learning of a smooth mapping from the coarser optimal gain surfaces in Fig.~\ref{fig:opt_cost_surface_all}, even in regions where there is a sharp gradient. This enables the trained models to serve as an efficient surrogate lookup, 
with each gain query resolved in approximately $\mathcal{O}(10^2)$~$\mu$s.

\vspace{-5pt}

\section{Conclusion} \label{sec:conclusion}

This paper addresses the problem of near-optimal gain selection for terminal angle-constrained guidance using two-phase Pure Proportional Navigation (2pPPN) guidance. By leveraging the flexibility in choosing the angular configuration at the end of the Orientation phase, a guidance effort minimization problem is formulated with navigation gains as the decision variables. As analytical solutions of this problem are not tractable, numerical solutions are obtained for various engagement geometries, showing that the optimal gain manifolds vary smoothly with the engagement configurations. A neural network regressor model is therefore trained to learn the mapping between the pair of initial and desired terminal engagement configurations and their corresponding optimal gains. Test results on the trained models predict near-optimal gains across the feasible engagement geometries with high accuracy level, thereby providing a single-shot near-optimal gain selector from the surrogate model-generated optimal gains manifold. Thus, the developed method facilitates a real-time optimal gains selection for a two-phase PPN-guided pursuer in an engagement with a stationary target point. 
\footnotesize
\bibliographystyle{ieeetr}
\bibliography{references}

\end{document}